\documentclass[12pt,aps,showpacs,superscriptaddress,footinbib,preprint,noshowpacs]{revtex4-1}
\usepackage{color}
\usepackage{graphicx}
\usepackage{amsmath}
\usepackage{comment}
\usepackage{amssymb}
\usepackage[normalem]{ulem}

\usepackage{subfigure}

\usepackage[table,xcdraw]{xcolor}

\usepackage{multirow}
\usepackage[version=3]{mhchem}
\newcommand{\etal}{\textit{et al.}}

\usepackage{float}
\usepackage{listings}
\usepackage{url}

\usepackage{breakurl}
\usepackage[breaklinks]{hyperref}
\usepackage{setspace}

\begin{document}

\title{Validating the Validation: Reanalyzing a large-scale comparison of Deep Learning and Machine Learning models for bioactivity prediction}

\begin{abstract}
Machine learning methods may have the potential to significantly accelerate drug discovery. However, the increasing rate of new methodological approaches being published in the literature raises the fundamental question of how models should be benchmarked and validated. We reanalyze the data generated by a recently published large-scale comparison of machine learning models for bioactivity prediction and arrive at a somewhat different conclusion. We show that the performance of support vector machines is competitive with that of deep learning methods.  Additionally, using a series of numerical experiments, we question the relevance of area under the receiver operating characteristic curve as a metric in virtual screening, and instead suggest that area under the precision-recall curve should be used in conjunction with the receiver operating characteristic. Our numerical experiments also highlight challenges in estimating the uncertainty in model performance via scaffold-split nested cross validation. 
\end{abstract}

\author{Matthew C. Robinson}
\affiliation{Department of Physics, J J Thomson Avenue, Cambridge. CB3 0HE}

\author{Robert C. Glen}
\affiliation{The Centre for Molecular Informatics, Department of Chemistry, University of Cambridge, Cambridge, UK}
\affiliation{Computational and Systems Medicine, Department of Surgery and Cancer, Faculty of Medicine, Imperial College London, UK}

\author{Alpha A. Lee}
\email{aal44@cam.ac.uk}
\affiliation{Department of Physics, J J Thomson Avenue, Cambridge. CB3 0HE}

\maketitle

\section{Introduction}
Computational approaches to drug discovery are often justified as necessary due to the prohibitive time and cost of experiments. Unfortunately, many papers fail to sufficiently prove that the proposed, novel techniques are actually an advance on current approaches when applied to realistic drug discovery programs. Models are often shown to work in situations differing greatly from reality, producing impressive metrics that differ greatly from the quantity of interest. It is then often the time and cost of properly implementing and testing these proposed techniques against existing methods that becomes prohibitive for the practitioner. There is also the significant opportunity cost if models prove to be inaccurate and misdirect resources.

These concerns are not new to the field of computational chemistry. Walters \cite{doi:10.1021/ci400197w}, Landrum and Stiefel \cite{landrum2012scientific}, and others have previously critiqued the state of the literature, even referring to many papers as ``advertisements". Furthermore, Nicholls has provided useful overviews of statistical techniques for uncertainty quantification and method comparison \cite{Nicholls2014, Nicholls2008, Nicholls2016}. Rarey and coworkers have recently provided an excellent critique of the importance of evaluating bias in data and the effects on deep learning algorithms in virtual screening \cite{sieg2019}. However, many authors still neglect the relevant issues, and results are still frequently reported without error bars, proper train and test set splitting, and easily usable code or data. 

Problems with validation are not unique to chemoinformatics or computational chemistry: numerous papers and manuscripts in the machine learning literature have been devoted to the proper evaluation of methods, with special concern to the applicability of statistical testing procedures for method comparison \cite{santafe2015dealing, derrac2011practical, dietterich1998approximate, demvsar2006statistical, japkowicz2011evaluating}. Recent reviews also provide background into procedures for hyperparameter optimization and model selection \cite{raschka2018model}. However, despite this work, researchers still find that new approaches frequently exploit bias in the training set, likely overfit to benchmark datasets \cite{DBLP:journals/corr/abs-1806-00451}, and even find that thousands of papers may be based on an initial result that was simply statistical noise \cite{border2019no}.

These errors are rampant for a number of reasons. One overarching issue is that the  literature has many differing suggestions and involves theoretical statistical ideas -- different metrics reward different aspects of the model, and commonly used metrics in machine learning do not necessarily reward models that are useful in drug design. Moreover, even when practitioners are interested in a thorough analysis of results, the task can be quite time intensive and costly. To properly test a new neural network architecture against older methods using several different random seeds, dataset splits, and learning rates might take on the order of 1000s of GPU hours and become a monetary concern. 

In this paper, we reanalyze the pioneering validation study by Mayr and coworkers \cite{mayr2018large}. Our study is made possible by their herculean effort in building a large-scale benchmarking study, as well as their generosity in making the code and data publicly available. The questions we will ask are: (1) Is one machine learning method significantly better than the rest, using metrics adopted by Mayr \etal{}? (2) Are the metrics adopted by Mayr \etal{} the most relevant to ligand-based bioactivity prediction? Our key conclusion is an alternative interpretation of their results that considers both statistical \textit{and} practical significance -- we argue that deep learning methods do not significantly outperform all competing methods. We also show, via a series of examples, that the precision-recall curve is relevant to ligand-based drug discovery and should be used in combination with the ROC-AUC metric. In reaching these conclusions, we also discuss and review issues of uncertainty and model comparison that are central to the field.  

The source code used for our reanalysis is available on GitHub \url{https://github.com/mc-robinson/validating_validation_supp_info}.   

\section{Study design of Mayr \etal{}}

Our study is motivated by the recent paper, entitled "Large-scale comparison of machine learning methods for drug target prediction on ChEMBL", by Mayr and coworkers. Realizing the recent success of deep learning in other fields and its introduction into drug-discovery \cite{goh2017deep, wu2018moleculenet}, Mayr and coworkers performed a large-scale evaluation of the method's success against other commonly-used machine learning methods in the drug discovery community. Their goal was to combat three common problems with model evaluation in chemical prediction: (1) a lack of large scale studies, (2) compound series bias in testing of drug-discovery algorithms, and (3) bias in hyperparameter selection. 

The Mayr \etal{} evaluation, based entirely on ligand-based approaches, had the explicit goal of comparing ``the performance of deep learning with that of other methods for drug target prediction." In pursuing this goal, the authors cited the relatively small number of assays in previous evaluation studies such as MoleculeNet \cite{wu2018moleculenet} and the need for larger scale evaluation. The authors believe that these small studies ``restrict the conclusions of the method comparisons to a certain subset of assays and underestimate the multitask learning effect in spite of the large
amount of data being available publicly." To correct this shortcoming, Mayr \etal{} extract data including roughly 456,000 compounds and over 1300 assays from ChEMBL \cite{bento2014chembl}. 

Notably, the ChEMBL data is quite heterogeneous. The diverse set of target classes includes ion channels, receptors, transporters, transcription factors, while the similarly diverse assay types include ADME, binding, functional, physiochemical, and toxicity assays. The number of compounds in each assay is also quite variable -- ranging from roughly one hundred compounds to over 30,000 in a given assay. In order to treat each problem as a separate binary classification procedure, the authors also develop a procedure to automatically convert the assay measurements to binary labels. Each assay is then treated as an individual classification problem. 

The compounds were then featurized using several different schemes including toxicophore features, semisparse features, depth first search features, and the popular (ECFP6) fingerprint \cite{rogers2010}. In our replication, we chose to use only the ECFP6 fingerprint because it could easily be constructed from the molecular data using the open-source program RDKit \cite{landrum2006rdkit}. 

\section{Is there a best model?}

In their study, Mayr \etal{} concluded that ``deep learning methods significantly outperform all competing methods." Much of this conclusion is based upon small $p$-values resulting from a Wilcoxon signed rank test used to quantify the differences in the average performance of the classifiers. For example, they report a $p$-value of $1.985 \times 10^{-7}$ for the alternate hypothesis that feedforward deep neural networks (FNN) has a higher area under the receiver operating curve (AUC-ROC) than support vector machines (SVM). For the alternative hypothesis that FNN outperform Random Forests (RF), the p-value is even more extreme ($8.491 \times 10^{-88}$). From such low $p$-values, one might be led to believe that FNN is the only algorithm worth trying in drug discovery. Yet, a closer look at the data reveals that this conclusion is clearly erroneous and obscures much of the variability from assay to assay.

\subsection{Are all assays created equal?}

To demonstrate the problems, we begin with an initial example of SVM and FNN performances using ECFP6 fingerprints. Table \ref{table:two_assays} shows AUC-ROC results from the FNN and SVM classifiers for two assays in the Mayr \etal{} dataset. Assay A is a functional assay consisting of a small number of samples. Each fold is heavily imbalanced and consists mostly of active compounds. Importantly, this is often the opposite imbalance one would observe in a real screen. As is expected with a small amount of highly imbalanced data, both the FNN and SVM classifier show highly variable results with very large confidence intervals. In fold 2, where only a single active compound is present in the test set, it is not even clear how to calculate the confidence intervals for AUC-ROC. The mean and standard error of the mean (SEM) are also calculated over the 3 folds, though this is slightly dangerous since it discards all of our knowledge of uncertainty in each fold. 

\begin{table*}[h!]
\centering
\begin{tabular}{lrccccclll}
 & \multicolumn{1}{l}{} & \multicolumn{1}{l}{} & \multicolumn{1}{l}{} & \multicolumn{1}{l}{} & \multicolumn{1}{l}{} & \multicolumn{1}{l}{} &  &  &  \\
 & \multicolumn{1}{c|}{\textbf{A: ChEMBL 1964055}} & \multicolumn{1}{c|}{Fold 1} & \multicolumn{1}{c|}{Fold 2} & \multicolumn{1}{c|}{Fold 3} & \multicolumn{1}{c|}{MEAN} & SEM &  &  &  \\ \cline{2-7}
 & \multicolumn{1}{r|}{\begin{tabular}[c]{@{}r@{}}FNN AUC-ROC\\ (95\% CI)\end{tabular}} & \multicolumn{1}{c|}{\cellcolor[HTML]{C0C0C0}\begin{tabular}[c]{@{}c@{}}0.44\\ (0.035, 0.94)\end{tabular}} & \multicolumn{1}{c|}{\cellcolor[HTML]{C0C0C0}\begin{tabular}[c]{@{}c@{}}0.62\\ (?, ?)*\end{tabular}} & \multicolumn{1}{c|}{\cellcolor[HTML]{C0C0C0}\begin{tabular}[c]{@{}c@{}}0.64\\ (0.34, 0.86)\end{tabular}} & \multicolumn{1}{c|}{\cellcolor[HTML]{EFEFEF}0.57} & \multicolumn{1}{c|}{\cellcolor[HTML]{EFEFEF}0.05} &  &  &  \\ \cline{2-7}
 & \multicolumn{1}{r|}{\begin{tabular}[c]{@{}r@{}}SVM AUC-ROC\\ (95\% CI)\end{tabular}} & \multicolumn{1}{c|}{\cellcolor[HTML]{C0C0C0}\begin{tabular}[c]{@{}c@{}}0.38\\ (0.02, 0.94)\end{tabular}} & \multicolumn{1}{c|}{\cellcolor[HTML]{C0C0C0}\begin{tabular}[c]{@{}c@{}}0.97\\ (?, ?)*\end{tabular}} & \multicolumn{1}{c|}{\cellcolor[HTML]{C0C0C0}\begin{tabular}[c]{@{}c@{}}0.68\\ (0.38, 0.88)\end{tabular}} & \multicolumn{1}{c|}{\cellcolor[HTML]{EFEFEF}0.67} & \multicolumn{1}{c|}{\cellcolor[HTML]{EFEFEF}0.14} &  &  &  \\ \cline{2-7}
 & \multicolumn{1}{r|}{\begin{tabular}[c]{@{}r@{}}Test Set Size\\ (Actives/ Inactives)\end{tabular}} & \multicolumn{1}{c|}{\cellcolor[HTML]{EFEFEF}\begin{tabular}[c]{@{}c@{}}35\\ (32/3)\end{tabular}} & \multicolumn{1}{c|}{\cellcolor[HTML]{EFEFEF}\begin{tabular}[c]{@{}c@{}}30\\ (29/1)\end{tabular}} & \multicolumn{1}{c|}{\cellcolor[HTML]{EFEFEF}\begin{tabular}[c]{@{}c@{}}35\\ (29/6)\end{tabular}} &  &  &  &  &  \\ \cline{3-5}
 & \multicolumn{1}{l}{} & \multicolumn{1}{l}{} & \multicolumn{1}{l}{} & \multicolumn{1}{l}{} & \multicolumn{1}{l}{} & \multicolumn{1}{l}{} &  &  &  \\
 & \multicolumn{1}{l|}{\textbf{B: ChEMBL 1794580}} & \multicolumn{1}{c|}{Fold 1} & \multicolumn{1}{c|}{Fold 2} & \multicolumn{1}{c|}{Fold 3} & \multicolumn{1}{c|}{Mean} & SEM &  &  &  \\ \cline{2-7}
 & \multicolumn{1}{r|}{\begin{tabular}[c]{@{}r@{}}FNN AUC-ROC\\ (95\% CI)\end{tabular}} & \multicolumn{1}{c|}{\cellcolor[HTML]{C0C0C0}\begin{tabular}[c]{@{}c@{}}0.889\\ (0.883, 0.895)\end{tabular}} & \multicolumn{1}{c|}{\cellcolor[HTML]{C0C0C0}\begin{tabular}[c]{@{}c@{}}0.905\\ (0.900, 0.910)\end{tabular}} & \multicolumn{1}{c|}{\cellcolor[HTML]{C0C0C0}\begin{tabular}[c]{@{}c@{}}0.906\\ (0.900, 0.911)\end{tabular}} & \multicolumn{1}{c|}{\cellcolor[HTML]{EFEFEF}0.900} & \multicolumn{1}{c|}{\cellcolor[HTML]{EFEFEF}0.005} &  &  &  \\ \cline{2-7}
 & \multicolumn{1}{r|}{\begin{tabular}[c]{@{}r@{}}SVM AUC-ROC\\ (95\% CI)\end{tabular}} & \multicolumn{1}{c|}{\cellcolor[HTML]{C0C0C0}\begin{tabular}[c]{@{}c@{}}0.926\\ (0.921, 0.931)\end{tabular}} & \multicolumn{1}{c|}{\cellcolor[HTML]{C0C0C0}\begin{tabular}[c]{@{}c@{}}0.926\\ (0.921, 0.930)\end{tabular}} & \multicolumn{1}{c|}{\cellcolor[HTML]{C0C0C0}\begin{tabular}[c]{@{}c@{}}0.934\\ (0.930, 0.939)\end{tabular}} & \multicolumn{1}{c|}{\cellcolor[HTML]{EFEFEF}0.929} & \multicolumn{1}{c|}{\cellcolor[HTML]{EFEFEF}0.002} &  &  &  \\ \cline{2-7}
 & \multicolumn{1}{r|}{\begin{tabular}[c]{@{}r@{}}Test Set Size\\ (Actives/ Inactives)\end{tabular}} & \multicolumn{1}{c|}{\cellcolor[HTML]{EFEFEF}\begin{tabular}[c]{@{}c@{}}19388\\ 5553/13855\end{tabular}} & \multicolumn{1}{c|}{\cellcolor[HTML]{EFEFEF}\begin{tabular}[c]{@{}c@{}}25165\\ 6918/18247\end{tabular}} & \multicolumn{1}{c|}{\cellcolor[HTML]{EFEFEF}\begin{tabular}[c]{@{}c@{}}19363\\ 5491/13872\end{tabular}} & \multicolumn{1}{l}{} & \multicolumn{1}{l}{} &  &  &  \\ \cline{3-5}
\end{tabular}
\caption{Two separate assays from the Mayr \etal{} data with the accompanying FNN and SVM prediction results. Confidence intervals for AUC-ROC are calculated through the Hanley-Mcneil method while the standard error of the mean (SEM) across folds is calculated in the standard fashion. * indicates that the confidence interval is unknown because it is not possible to calculate the effective degrees of freedom in the usual way, when only one sample is given from the positive class. }
\label{table:two_assays}
\end{table*}

\begin{figure}
\centering
\subfigure[]{\includegraphics[scale=0.6]{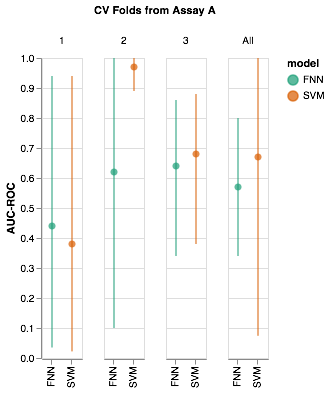}}
\subfigure[]{\includegraphics[scale=0.6]{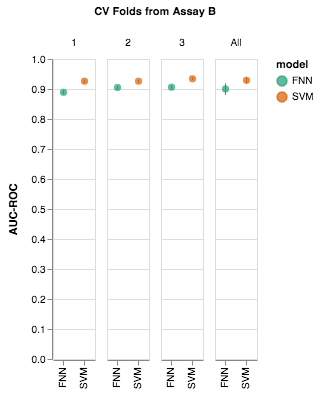}}
\caption{Figures displaying the information contained in Table \ref{table:two_assays}. In the case of fold 2 for assay A, 95\% confidence intervals are calculated based on the assumption that the confidence interval can be derived from the Hanley-Mcneil standard error; this is a large assumption. For the mean values across all folds, a $t$-distribution with two degrees of freedom is used to calclulate 95\% confidence intervals for the mean AUC-ROC across the 3 folds. }
\label{fig:two_assays_bar_charts}
\end{figure}

In contrast, assay B is a functional assay with large samples and imbalances that more closely resemble those typically seen in the literature. Performance is quite good, with the SVM classifier outperforming the FNN classifier on each fold. Furthermore, the confidence intervals for each AUC-ROC value are quite small. Again the mean and SEM are calculated across the folds for each classifier. Additionally, Figure \ref{fig:two_assays_bar_charts} gives a visual representation of the performances for assay A and assay B.

One would likely agree that Figure \ref{fig:two_assays_bar_charts} shows a striking difference between the results of the two assays. While the results of assay A for FNN and SVM are extremely noisy and raise many questions, assay B shows a well-defined difference in the performance of the two algorithms, even relative to the noise levels of the measurements. Though not a formal analysis, due to the presence of noise, one would likely consider the difference in mean performances on assay A, $0.67 - 0.57 = 0.10$, to be much less meaningful than the difference in mean performances on assay B, $0.929 - 0.900 = 0.029$. To most practitioners, comparative performances on assay B would give much more evidence to SVM outperforming FNN than the comparative performances on assay A, even though it is lower in magnitude. More formally, the effect size in assay B (4.40) is approximately eight times the size of the effect size in assay A (0.55). 

Nevertheless, the Wilcoxon ranked sign test used by Mayr \etal{} treats these numbers as commensurate, and assumes that the larger magnitude difference of mean AUC-ROC values in assay A should carry more weight than the smaller magnitude difference of mean AUC-ROC values in assay B. This is not necessarily true. Firstly, effect size, which measures the magnitude of the difference relative to the uncertainty, is more important than pure magnitude differences. Secondly, differences in AUC/probability space are not straightforward: $p=0.01$ and $p=1\times10^{-6}$ have a smaller difference in absolute magnitude than $p=0.51$ and $p=0.52$; however, the difference between one in 100 and one in a million is likely much more important than the difference between 51\% vs 52\%. Lastly, these concerns aside, assuming commensurate results was already problematic given the heterogeneity of the assay types, changing sample sizes, varying imbalances, and diverse target classes. 

\subsection{To win or not to win, this is the question}

Having realized that the Wilcoxon ranked sign test is inappropriate, we turn to the sign test as perhaps the most appropriate procedure. The sign test essentially counts the proportion of ``wins" for a given algorithm over another on all of the datasets. As with many of these tests, the null hypothesis of the sign test is that the two models show the same AUC-ROC performance on the datasets. Assuming this null hypothesis, the algorithm displaying the better performance on a given assay should be determined by a coin-flip. Therefore, given $N$ assays, we expect each classifier to win on approximately $N/2$ assays. There are, of course, still problems with the sign test. First, the test still discards most of the uncertainty information. Secondly, the test still counts the assays in Table \ref{table:two_assays} and Figure \ref{fig:two_assays_bar_charts} of equal weight, which is better than in a rank test, but still suboptimal. Additionally, due to the lack of parametric assumptions, the sign test has low power, meaning that it often fails to detect a statistically significant difference in algorithms when one exists. 

Using the sign-test we calculated 95\% Wilson score intervals for the sign-test statistic for the alternative hypothesis that FNN has better AUC-ROC performance than SVM, the second best performing classifier according to Mayr \etal{}. Using all folds in the analysis (since each is indeed an independent test set) gives an interval of (0.502, 0.534), while only comparing the mean AUC values per assay gives a confidence interval of (0.501, 0.564). While both of these tests are narrowly significant at the $\alpha=0.05$ level (intervals do not include $0.5$), it is worth examining the practical meaning of these results. According to the statistic, our data is compatible with an FNN classifier beating an SVM classifier on 50\% to 56\% of the assays, Thus, if one were to conclude that only an FNN classifier is worth trying, the user would be failing to use a better classifier almost 50\% of the time! And this is in the case of a two classifier comparison. Considering all the classifiers, FNN and SVM both perform the best in 24\% of the assays, while every other classifier considered by Mayr \etal{} is the best performing classifier on at least 5\% of the assays (Table \ref{win_table} shows a breakdown of wins). Clearly, some of these results are just noise due to small assay sizes; however, it indicates that classifier performance is likely assay dependent, and one should try multiple classifiers for a given problem.

\begin{table*}[]
\centering 
\begin{tabular}{|l|l|l|l|l|l|l|l|l|l|}
\hline
Method                                                                                        & SVM                     & FNN                     & GC                      & NB                      & LSTM                    & RF                      & Weave                   & KNN                     & Tie                     \\ \hline
\multirow{2}{*}{\begin{tabular}[c]{@{}l@{}}\% of folds that the method \\ is the best performing method\end{tabular}} & \multirow{2}{*}{24.6\%} & \multirow{2}{*}{24.6\%} & \multirow{2}{*}{9.99\%} & \multirow{2}{*}{9.69\%} & \multirow{2}{*}{7.91\%} & \multirow{2}{*}{7.71\%} & \multirow{2}{*}{7.51\%} & \multirow{2}{*}{5.73\%} & \multirow{2}{*}{2.34\%} \\
                                                                                                 &                         &                         &                         &                         &                         &                         &                         &                         &                         \\ \hline
\end{tabular}
                                                                                                 
\caption{The percent of folds, across all assays, that a method is the best performing method. SVM and FNN are the clearly the best performing methods, and it is noteworthy that SVM outperforms deep learning methods such as GC, LSTM and Weave.  }
\label{win_table}
\end{table*}

\begin{figure}
\centering
\includegraphics[width=4in]{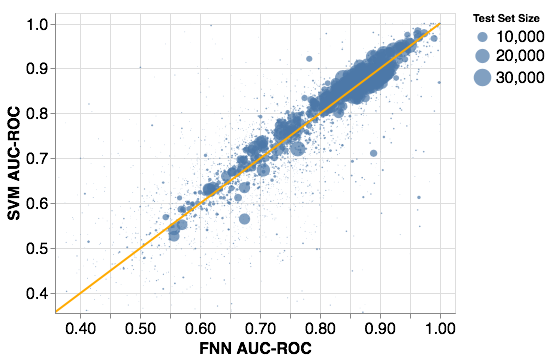}
\caption{A comparison of FNN and SVM AUC-ROC performance on all test folds. The orange line indicates the identity line of slope 1, while the dot size indicates the size of the test set.} 
\label{fig:FNN_vs_SVM_AUC}
\end{figure}

\begin{figure}[!ht]
\centering
\includegraphics[width=4in]{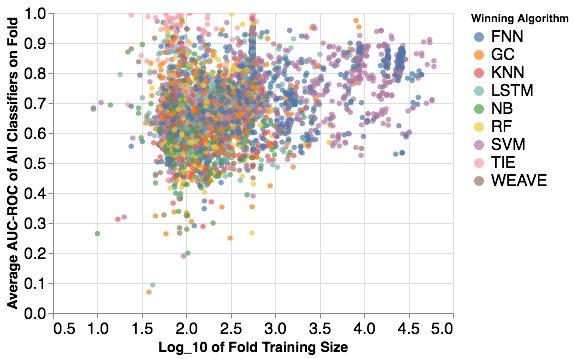}
\caption{The best performing algorithm (indicated by color) in terms of AUC-ROC for all test folds as the training size increases.} 
\label{fig:AUC_WINNERS}
\end{figure}

The above considerations are illustrated by the data in Figures \ref{fig:FNN_vs_SVM_AUC} and \ref{fig:AUC_WINNERS}. Figure \ref{fig:FNN_vs_SVM_AUC} shows that the FNN and SVM performance is almost identical for large datasets, but the difference between the performances varies quite sporadically for assays with fewer compounds (the smaller points in the figure). Additionally, Figure \ref{fig:AUC_WINNERS} shows the best performing algorithm for each independent test fold; we also plot the other algorithms that Mayr \etal{} considered, namely random forest (RF), k-nearest neighbours (kNN), Graph Convolutional neural networks (GC), Weave, and Long Short-Term Memory networks with SMILES input (LSTM). As one can see, the results are quite varied for smaller assays, and the best performing algorithm is largely dataset dependent. Much of this variation is due to the 3-fold CV procedure of Mayr \etal{} that is quite susceptible to large variations because of the small dataset size. However, as the training size increases, the deep learning and SVM algorithms dominate. Interestingly, among all datasets with greater than 1,000 compounds in the test set, SVM performance is better than FNN performance on 62.5\% of assays, which is counter to the usual wisdom that deep learning approaches beat SVMs in large assays. Notably, GC, LSTM, and Weave, show the best performance on only a small number of large assays, casting doubt on their utility over a standard FNN or SVM. With all of these observations, it should be noted that the results could be due to sub-optimal hyperparameter optimization -- and perhaps some of these models can achieve state-of-the-art performance in the hands of expert users. However, hyperparameter optimisation can take a considerable amount of time and computing resources.

\begin{figure}
\centering
\includegraphics[width=7in]{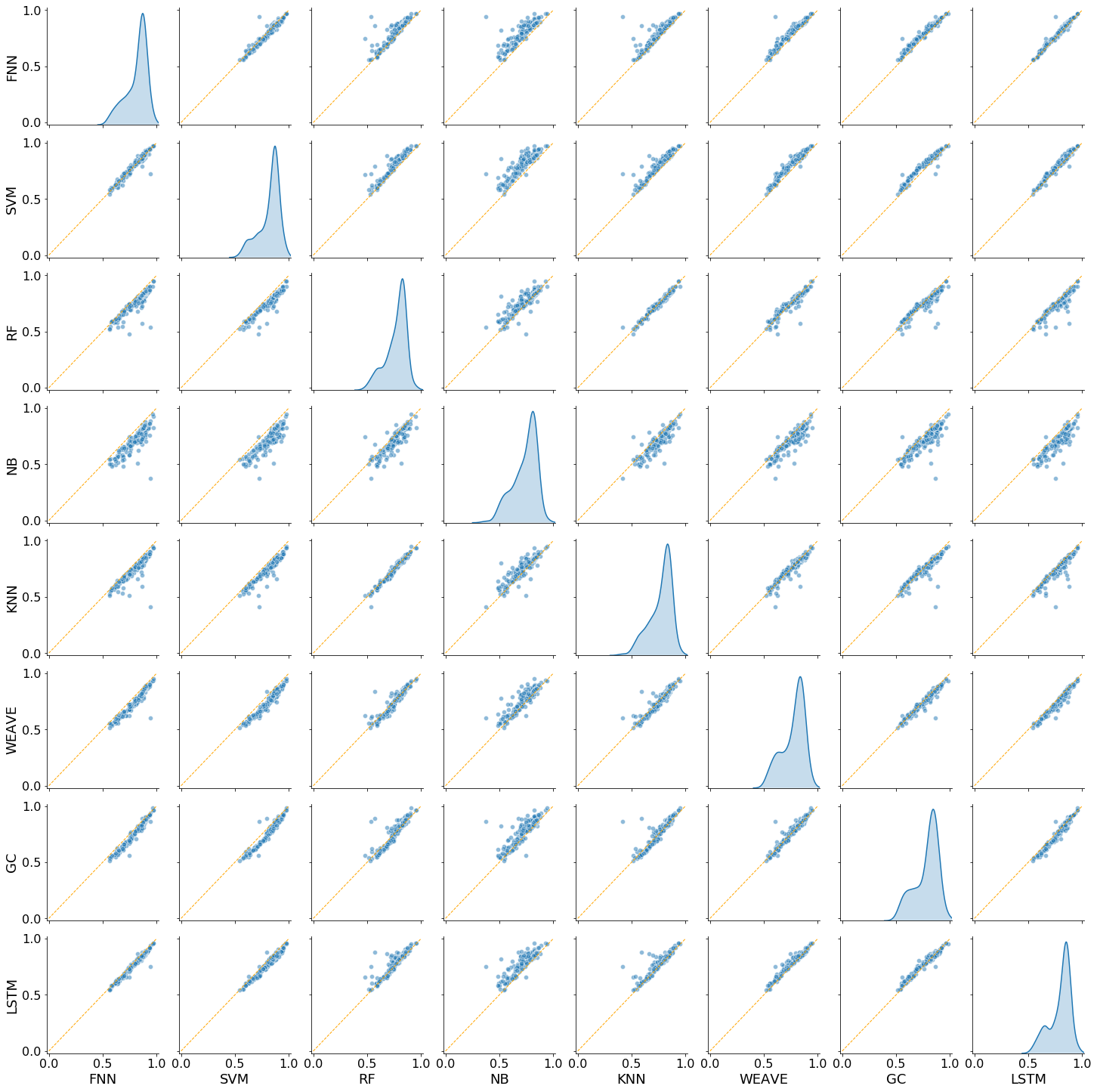}
\caption{The correlation in mean AUC-ROC performance for all models with more than 1000 samples in the test set on all assays. The orange line indicates the identity line of slope 1, while the dot size indicates the size of the test set. The density plots on the diagonal represent the distribution of mean AUC-ROC scores for the given classifier.} 
\label{fig:large_pairwise}
\end{figure}

Additionally, the correlation between the mean AUC-ROC performances for all models is shown in Figure \ref{fig:large_pairwise} for all assays with more than 1000 test set samples on average over the 3 folds. As can be seen, most of the deep learning models perform quite similarly, with NB, KNN, and Weave seeming to show the worst performance. The figure showing how well the models correlate on assays of all sizes is also included in the supporting information. Unfortunately, it is tough to make inferences regarding the relative performance in small sizes due to the inherent noise of the datasets and 3-fold CV procedure. 

Taking all of the above into consideration, it appears that the FNN and SVM models are the best performing models, especially in the case of large datasets. In small datasets, NB, KNN, and RF can often still perform competitively. It is also unclear how well the frequently used gradient-boosted decision tree algorithm would compare, since it was not included in the study. The Mayr \etal{} data contains quite a lot of information and we provide it and our code online for all who wish to further analyze it.

\section{What performance metric do we need?}

Having reevaluated different machine learning methods using the metric suggested by Mayr \etal{}  -- ROC-AUC -- we now turn to consider the more general question of what performance metric is most closely correlated to practical success in drug discovery. 

\subsection{What does ROC-AUC measure?}

Typical machine learning performance measures such as accuracy, true positive rate (TPR), false positive rate (FPR), precision, specificity, are combinations of the entries of the confusion matrix. In the virtual screening literature, researchers frequently use the area under the curve (AUC) of the receiver operating characteristic curve (ROC), which plots TPR vs FPR. The AUC-ROC is not based on a single threshold, and instead gives an indication of classifier performance over a range of varying classification thresholds. In this way, the AUC-ROC captures how well a classifier discriminates between the classes of interest. Conveniently, the AUC-ROC can also be interpreted as the probability that a randomly chosen member of the positive class will be correctly ranked before a randomly chosen member of the negative class. Therefore, an AUC-ROC of $1.0$ indicates perfect discrimination between classes, and an  AUC-ROC of $0.5$ indicates random guessing. 

While the AUC-ROC is more robust than metrics such as accuracy in cases of class imbalance, it is still not without criticism. These critiques, popularized by Hand and coworkers \cite{hand2009measuring}, are perhaps best understood if one interprets the ROC-AUC as the expected TPR averaged over all classification thresholds (false positive rates). Therefore, if two ROC curves cross, the AUC of one curve may be higher even if it performs much worse (has a lower TPR) over the region containing the classification thresholds of interest. Additionally, Hand raises concerns about the ``incoherence" of AUC-ROC, since the measure ignores relative cost concerns of each threshold when simply taking an expected value over all such decision thresholds (FPR from zero to one). 

The critique of AUC-ROC most widely seen in drug discovery is that it does not account for the ``early behavior" of a classifier. Since the purpose of virtual screening procedures is often to rank the compounds by likely activity and avoid experimentally screening an intractable number of compounds, the classifier is only useful if active compounds are ranked at the top of the list and prioritized for actual screening. Unfortunately, the AUC-ROC does not take into account this early performance and only measures average discrimination performance. 

As a result of these shortcomings, many alternative methods including BedROC and RIE have been proposed, as described in \cite{Nicholls2008}. However, these methods are sensitive to a tunable parameter and are not as interpretable as a metric as AUC-ROC. In drug discovery, the enrichment factor is often used to quantify this early behavior, which describes how many of the total actives are found in the top X\% of ranked compounds. Unfortunately, this metric can be quite noisy and is sensitive to both the chosen percentage and specific ordering of compounds at the top of the list. Alternatives such as ROC enrichment, which instead uses the fraction of inactives and is related to the AUC-ROC, have also been proposed \cite{Nicholls2008}.

Instead of focusing on specific drug discovery metrics, we propose that the widely used area under the precision-recall curve (AUC-PRC) may serve as an important complement to the AUC-ROC in chemical applications. To illustrate why AUC-PRC is more appropriate than AUC-ROC, we describe below a series of numerical simulations. Our simulations build on the results of Saito and Rehmsmeier \cite{saito2015precision}.

\subsection{Precision-recall should be used in conjunction with AUC-ROC}

\begin{figure}[!ht]
\centering
\subfigure[]{\includegraphics[width=2.5in]{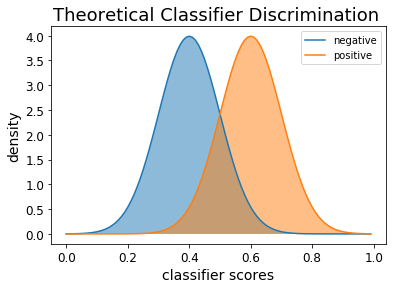}}
\subfigure[]{\includegraphics[width=3.8in]{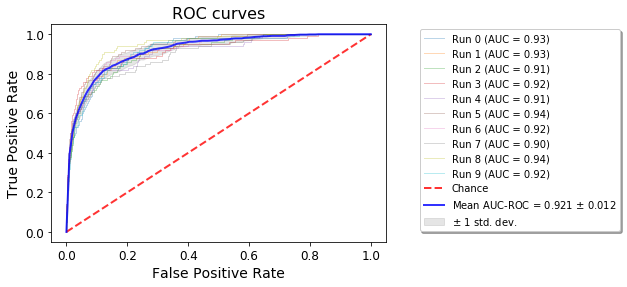}}
\subfigure[]{\includegraphics[width=3.5in]{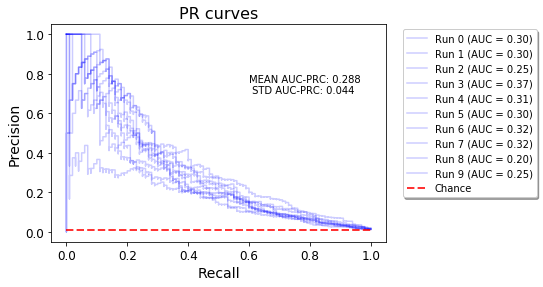}}
\subfigure[]{ \includegraphics[width=2.8in]{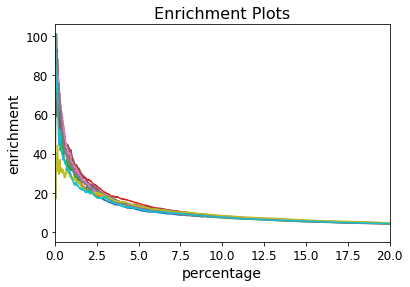}}
\caption{(a) Distributions of classification scores for a theoretical classifier of positive and negative examples. The negative scores follow a $\mathcal{N}(\mu=0.4,\,\sigma=0.1)$ distribution, while the positive scores follow a $\mathcal{N}(\mu=0.6,\,\sigma=0.1)$. Adapted from the simulations in \cite{saito2015precision}. (b) ROC, (c) PRC, and (d) enrichment factor curves for predictions from the theoretical classifier shown in (a). The curves result from 10 runs of a simulation with large class imbalance ($\sim 1\%$ actives)}
\label{fig:my_prc_no_early_simulation results}
\end{figure}

We first consider a theoretical classifier of positive and negative examples, shown in Figure \ref{fig:my_prc_no_early_simulation results}a. This theoretical classifier shows decent discrimination between the two classes, as shown by the separation of the two normal distributions. 

Because this hypothetical classifier assigns scores to each class based on well known distributions, we can computationally take samples from each distribution. In our example, we take $N_+=100$ samples from the positive class and $N_-=10,000$ samples from the negative class in order to mimic a 1\% hit rate of actives that might be observed in a virtual screen. After repeating this experiment ten times, we then plot ROC curves, PRC curves, and percent enrichment factor curves for each round of the simulation, as shown in figure \ref{fig:my_prc_no_early_simulation results}b-d.

The first thing to notice in these plots is the large discrepancy between the average AUC-ROC and AUC-PRC scores. This difference results from the large number of false positives which cause the precision-recall scores to be quite low. Importantly, one may wrongly conclude that the performance of the classifier is almost perfect by merely observing the ROC plot, while observing the PRC plot indicates poor precision. 

\begin{figure}
\centering
\subfigure[]{\includegraphics[width=2.5in]{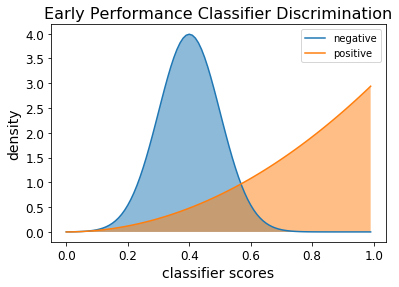}}
\subfigure[]{\includegraphics[width=3.8in]{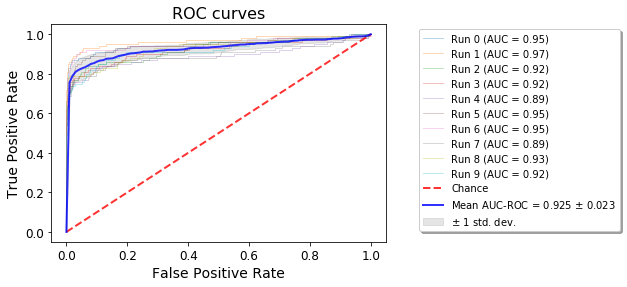}}
\subfigure[]{\includegraphics[width=3.6in]{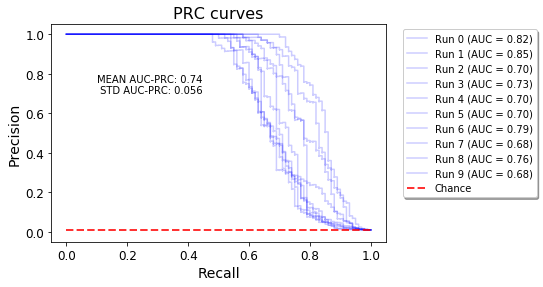}}
\subfigure[]{ \includegraphics[width=2.6in]{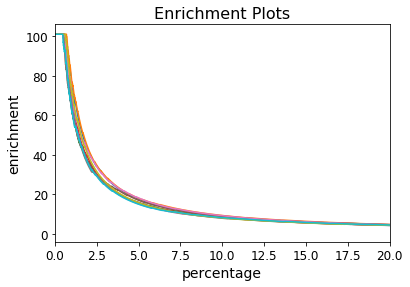}}
\caption{(a)Distributions of classification scores for a theoretical classifier of positive and negative examples with improved early performance. The negative scores follow a $\mathcal{N}(\mu=0.4,\,\sigma=0.1)$ distribution as before in Figure \ref{fig:my_prc_no_early_simulation results}a, while the positive scores follow a $B(a=3,b=1)$ distribution. (b) ROC, (c) PRC, and (d) enrichment factor curves for predictions from the theoretical classifier shown in (a). The curves result from 10 runs of a simulation with large class imbalance ($\sim 1\%$ actives)}
\label{fig:my_prc_early_simulation results}
\end{figure}

We next consider an alternate theoretical classifier with improved early performance in Figure \ref{fig:my_prc_early_simulation results}a. In this case, the negative samples are drawn from the same normal distribution as in the aforementioned simulation, while the positive samples are drawn from a Beta distribution to bias the results towards improved early retrieval of actives. 

The same simulation procedure is repeated with the same class imbalance, and the results of the simulation are shown in Figure \ref{fig:my_prc_early_simulation results}b-d. Notably, the average AUC-ROC is almost the same as in the previous simulation but the average AUC-PRC and enrichment factors show a marked increase. Therefore, we observe that the precision-recall curve better accounts for the desired early performance behavior in drug-discovery applications. To show that this is not merely due to intricacies of our simulation setup, we replicate the simulations of Saito and Rehmsmeier, which show the same effect, in the supporting information.

\begin{figure}[!h]
\centering
\includegraphics[width=6in]{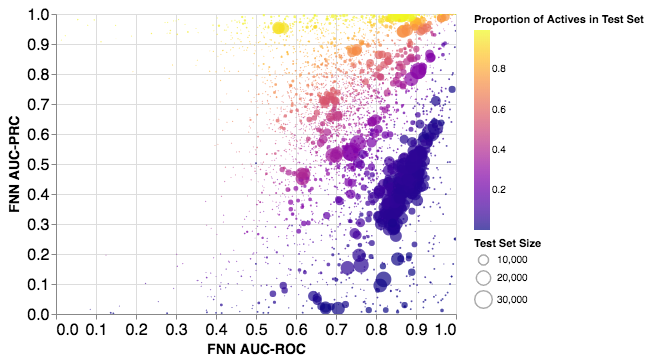}
\caption{The relationship between AUC-ROC scores and AUC-PRC scores for all test folds for the FNN model. Note the limited range of AUC-ROC scores in comparison to the AUC-PRC scores, which vary considerably. We can only compare ROC and PRC for FNN because PRC values were not reported for the other methods.}
\label{fig:PRC_vs_ROC}
\end{figure}

As an example of the utility of the AUC-PRC score, we plot the AUC-ROC and AUC-PRC performances of FNN on all 1310 assays. As one can see in Figure \ref{fig:PRC_vs_ROC}, the two metrics are not necessarily well correlated, and thus indicate large class imbalances. Additionally, observing that the AUC-PRC was often large when the AUC-ROC indicated mediocre performance alerted us to the fact that many of the assays show the opposite class skew we would expect from virtual screening assays. Importantly, a majority of the labeled compounds were active in a large number of assays (in 165 of the 1310 assays, there is at least one test fold consisting of 90\% or greater actives), even though virtual screening applications usually involve situations where the inactives far outnumber the active hits (in nearly all collected and literature published datasets, molecular diversity is generally greater and not focused on an active series or related molecules). 

\section{How should models be trained and tested?}

Another major issue in the chemical machine learning literature that Mayr and coworkers hoped to address was the way models are trained and tested. They combined two innovative methodologies -- nested cross validation and scaffold splitting. We will summarise those methodologies below, and then discuss the tradeoffs.   

\subsection{Cross-validation and scaffold splitting}

Machine learning models often contain many hyperparameters. Cross-validation (CV) is a strategy that enables the user to tune those hyperparameters without overfitting. The nested CV protocol consists of an outer and inner CV loop. The entire procedure is perhaps best understand as a simple $k$-fold CV procedure, in which the holdout test set is one of $n$ distinct folds. The $n \times k$ nested CV setup thus consists of $n$ different simple $k$-fold CV procedures for model selection, followed by model evaluation on the $n$ distinct testing folds. Accordingly, the average of the performance on the $n$ testing folds provides almost a completely unbiased estimate of the true generalization performance \cite{varma2006bias, raschka2018model}.

However, cross validation alone is insufficient to estimate the true performance of models. Compounds in chemical datasets are often centered around easily synthesized scaffolds, which are then modified by adding functional groups. Therefore, certain machine learning algorithms may just memorize properties of certain scaffolds and fail to generalize to new chemicals. Thus, if similar compounds are contained in both the training and test sets, we expect that estimates of machine learning performance would overestimate the true generalization performance on new compounds. 

To counteract this problem, it is now popular to perform ``scaffold splitting", where compounds are split into subsets based on their two-dimensional structural framework. An implementation is included in the DeepChem package \cite{deepchem}, and the results of random splitting have been explored both in the MoleculeNet benchmark paper, wherein Wu and coworkers reported larger differences between training and testing performance with scaffold splitting than with random splitting, as is expected. 

\subsection{Average and outliers}

We return now to examining the specific model evaluation and comparison approaches found in Mayr \etal{}. Generally, their approach involves performing the aforementioned $3 \times 2$ nested cluster-cross-validation model evaluation procedure on all 1310 assays. Importantly, due to the setup of their clustering approach, the number of compounds in each fold is not the same. Furthermore, the ratio of active to inactive compounds in each fold may be quite different. For example, consider the FNN results shown in Table \ref{table:bad_assay} on a particularly troublesome assay CHEMBL1243971, which measures the inhibition of the PI4-kinase beta subunit. This assay is one of the smallest in the entire dataset, and includes folds that are heavily imbalanced in terms of size. In order to compare the performance of FNN to other models on this assay, the mean and standard deviation AUC-ROC scores over all 3 folds were calculated by Mayr \etal{}. However, this averaging completely discards the inherent uncertainty of each independent test fold, which can be useful information. 

\begin{table}[!ht]
\centering
\begin{tabular}{lccc}
\multicolumn{1}{c}{} & \multicolumn{1}{c|}{Fold 1} & \multicolumn{1}{c|}{Fold 2} & Fold 3 \\ \cline{2-4} 
\multicolumn{1}{c|}{AUC-ROC} & \multicolumn{1}{c|}{0.69} & \multicolumn{1}{c|}{0.00} & \multicolumn{1}{c|}{0.56} \\ \hline
\multicolumn{1}{l|}{\begin{tabular}[c]{@{}l@{}}Test Set Number\\ Actives/Inactives\end{tabular}} & \multicolumn{1}{c|}{18/18} & \multicolumn{1}{c|}{2/1} & \multicolumn{1}{c|}{3/3} \\ \cline{2-4} 
 & \multicolumn{1}{l}{} & \multicolumn{1}{l}{} & \multicolumn{1}{l}{}
\end{tabular}
\caption{The results of an FNN deep learning model on the ChEMBL 1243971 assay. The AUC-ROC scores for the three disjoint test folds are reported.}
\label{table:bad_assay}
\end{table}

Taking fold 2 as an example, we can take the approach of \cite{hanley1982meaning} and recognize that the AUC-ROC, which is again the probability that a randomly chosen positive sample is correctly ranked higher than a randomly chosen negative sample, is equivalent to the value of the Wilcoxon-Mann-Whitney statistic. Doing this calculation, we find that a classifier can achieve three possible values of AUC-ROC with two active and one inactive compound in the test set, AUC-ROC $=0.0$, AUC-ROC $=0.5$, and AUC-ROC $=1.0$ . Therefore, a completely random classifier, would achieve a mean AUC of $0.5$ on fold 2 with a standard deviation of $0.41$. Thus, most confidence intervals of interest would reach far outside the possible range of AUC values, $[0,1]$, and any result on this fold is essentially meaningless. Even the larger, more balanced, fold 1 has a relatively large 95\% confidence interval of AUC-ROC $\in [0.49, 0.84]$ using the approximation in \cite{Nicholls2014} and thus cannot be rejected as a random classifier (AUC-ROC $=0.5$) at the $\alpha=0.05$ significance threshold. Importantly, since averaging treats all folds the same, the result on fold 1 is treated as of equal importance to the meaningless result on fold 2. Furthermore, one outlier fold performance can significantly affect the average AUC on a given assay. Thus, the use of unequal fold sizes coupled with mean AUC scores renders interpretation challenging.

\subsection{Cross-validation underestimates error}

Leaving the concern of variable amounts of data for each fold aside, a fundamental question is whether cross-validation provides an accurate measure of error.  Varoquaux previously performed extensive simulations to show that the standard error across cross-validation folds considerably underestimates the actual error \cite{varoquaux2018cross}. To understand the extent of this error, we adapted his simulation procedures focusing on prediction accuracy to measure the systematic errors in AUC-ROC estimation from cross-validation. All of our code for these simulations has been made available in the supplementary information.

\begin{figure}[!ht]
\centering
\includegraphics[width=5in]{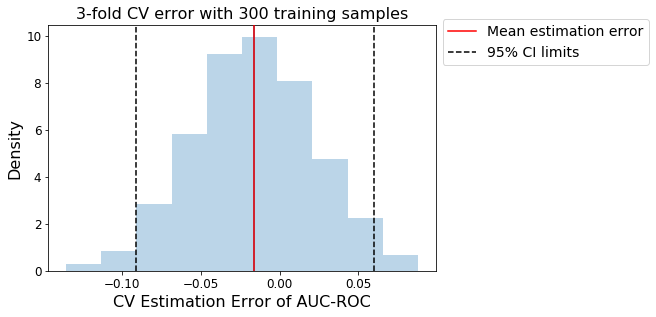}
\caption{The distribution of the errors when estimating true generalization performance from 3-fold CV for $N_{train}=300$ training samples. The dotted black lines indicate the $2.5$ and $97.5$ percentiles, thus denoting the boundaries of a 95\% confidence interval.}
\label{fig:gael_300_roc}
\end{figure}

We begin by constructing an artificial high-dimensional dataset of two relatively well separated Gaussian distributions. The separation of the distributions is adjusted such that a Linear SVM classifier will achieve an AUC-ROC $=0.75$ on the data. A training set of size $N_{train}$ is then drawn from the artificial two-class dataset. The Linear SVM is trained on these $N_{train}$ samples then deployed on a test set of size 10,000 samples, which gives an estimate of the true generalization performance. This true generalization performance is compared to the mean performance of a 3-fold CV procedure on the original $N_{train}$ training samples. Thus, we can compare the classifier's true generalization AUC-ROC performance to the estimate of that performance from cross-validation. Figure \ref{fig:gael_300_roc} shows the results from $1,000$ runs of the simulation using $N_{train}=300$ training samples. As can be seen, the CV estimates of performance are still frequently off by over 0.05 in AUC-ROC space. Moreover, the distribution of errors is asymmetric, as is to be expected for AUC-ROC. It should be noted that these large errors result even in the case of an artificial dataset of well-behaved distributions and equal sized folds with small class imbalance.

In addition to measuring CV errors in estimation of generalization performance, we can use these simulation results to measure how well the standard error of the mean (SEM) represents the error of the CV procedure. To measure this, we construct 95\% confidence intervals centered around the mean 3-fold CV performance. Theoretically, 95\% of the confidence intervals constructed in this fashion should contain the true generalization performance if the cross-validation procedure is unbiased. However, we find that the confidence interval coverage is only 79.7\% for $N_{train}=300$, even when using the extremely generous bounds to the $t$-distribution with 2 degrees of freedom. The coverage is similarly poor for other sizes of $N_{train}$, and would be much worse if one naively used the $\pm 2 \times (SEM)$ rule to construct confidence intervals. These results unfortunately indicate that the error bars on cross validation means are often too optimistic. This bias results from the correlation of training data across folds, thus violating independence assumptions. These simulation results cast doubt on the ability of cross validation procedures with small sample sizes to accurately reflect the generalization performance of a classifier with appropriate uncertainty.

\section{Conclusion}

We build on the recent large-scale benchmarking study by Mayr and coworkers and reanalysed the reported performance data of different machine learning models, arriving at a different conclusion to Mayr and coworkers. We show that support vector machines achieve competitive performance with feed-forward deep neural networks. Moreover, we show, via numerical simulations, that the area under the precision-recall curve is often more informative than the area under the receiver operating characterise curve in terms of assessing the performance of machine learning models in contexts relevant to drug discovery. We also highlight challenges in interpreting scaffold splitting cross validation results.

\begin{acknowledgements}
AAL acknowledges the support of the Winton Programme for the Physics of Sustainability. 
\end{acknowledgements}

\bibliography{references.bib}

\end{document}